\documentclass[10pt,twocolumn,letterpaper]{article}

\usepackage[pagenumbers]{cvpr} 

\usepackage[dvipsnames]{xcolor}

\usepackage{graphicx}
\usepackage{amsmath}
\usepackage{amssymb}
\usepackage{booktabs}
\usepackage{cancel}
\usepackage{pifont}
\usepackage{subfloat}
\usepackage{stfloats}
\usepackage{float}
\usepackage{booktabs}
\usepackage{threeparttable}
\usepackage[accsupp]{axessibility} 
\definecolor{cvprblue}{rgb}{0.21,0.49,0.74}
\usepackage[pagebackref,breaklinks,colorlinks,citecolor=cvprblue]{hyperref}

\def\paperID{13265} 
\def\confName{CVPR}
\def\confYear{2024}

\title{Salience DETR: Enhancing Detection Transformer \\
with Hierarchical Salience Filtering Refinement}

\author{Xiuquan Hou$^1$, Meiqin Liu$^{1,2,}$\thanks{Corresponding author}~, Senlin Zhang$^2$, Ping Wei$^1$, Badong Chen$^1$\\
$^1$National Key Laboratory of Human-Machine Hybrid Augmented Intelligence, \\
National Engineering Research Center for Visual Information and Applications, \\
and Institute of Artificial Intelligence and Robotics, Xi'an Jiaotong University, Xi'an, China\\
$^2$College of Electrical Engineering, Zhejiang University, Hangzhou, China\\
{\tt\small xiuqhou@stu.xjtu.edu.cn, liumeiqin@zju.edu.cn, slzhang@zju.edu.cn,}\\
{\tt\small pingwei@mail.xjtu.edu.cn, chenbd@mail.xjtu.edu.cn}
}

\begin{document}
\maketitle
\begin{abstract}
    DETR-like methods have significantly increased detection performance in an end-to-end manner. The mainstream two-stage frameworks of them perform dense self-attention and select a fraction of queries for sparse cross-attention, which is proven effective for improving performance but also introduces a heavy computational burden and high dependence on stable query selection. This paper demonstrates that suboptimal two-stage selection strategies result in scale bias and redundancy due to the mismatch between selected queries and objects in two-stage initialization. To address these issues, we propose hierarchical salience filtering refinement, which performs transformer encoding only on filtered discriminative queries, for a better trade-off between computational efficiency and precision. The filtering process overcomes scale bias through a novel scale-independent salience supervision. To compensate for the semantic misalignment among queries, we introduce elaborate query refinement modules for stable two-stage initialization. Based on above improvements, the proposed Salience DETR achieves significant improvements of +4.0\% AP, +0.2\% AP, +4.4\% AP on three challenging task-specific detection datasets, as well as 49.2\% AP on COCO 2017 with less FLOPs. The code is available at \href{https://github.com/xiuqhou/Salience-DETR}{https://github.com/xiuqhou/Salience-DETR}.
\end{abstract}

\section{Introduction}
\label{sec:intro}
Object detection is a fundamental task in computer vision with numerous practical applications. Despite the significant advancements made by convolutional detectors \cite{ren2015faster, redmon2016you, liu2016ssd, cai2018cascade} in recent decades, they are still limited by manually-designed components such as non-maximum suppression \cite{carion2020end}. With the recent advent of DEtection TRansformer (DETR) \cite{carion2020end}, end-to-end transformer-based detectors have shown remarkable performance improvement in the COCO challenge \cite{ye2023cascade, zong2023detrs}.
\begin{figure}[!tbp]
  \centering
  \includegraphics[width=0.45\textwidth]{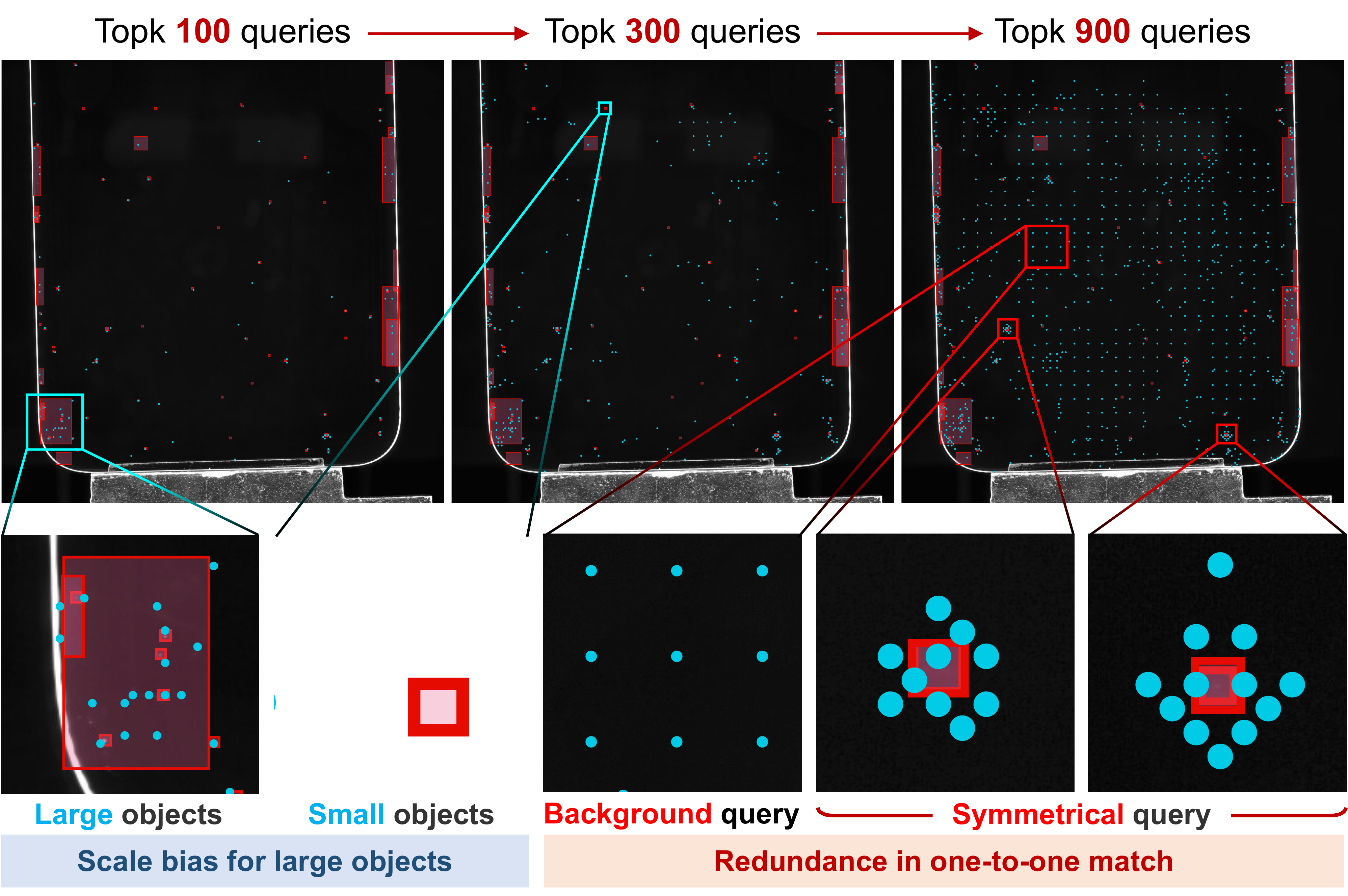}
  \caption{Visualization of \textit{selected queries in two-stage initialization}. Queries and object annotations are denoted in \textcolor[RGB]{0,203,230}{\textbf{points}} and \textcolor[RGB]{231,13,5}{\textbf{bounding boxes}} respectively. The selection results illustrate scale bias and redundancy despite one-to-one Hungarian matching.}
  \label{fig:query_visualization}
\end{figure}

Among the large number of variants of DETR, the latest high-performance frameworks follow a two-stage pipeline that performs dense self-attention in the encoder and selects sparse queries for cross attention in the decoder \cite{zhu2020deformable, liu2021dab, li2022dn, zhang2022dino, chen2023group, ye2023cascade, zong2023detrs}. This does improve the detection performance but also results in increased computation and the requirement for stable two-stage query initialization \cite{zheng2023less}. As shown in \Cref{fig:query_visualization}, we observe that in task-specific detection scenarios involving weak objects (\eg small-scale objects affected by scattering and low contrast \cite{hou2023canet}), existing two-stage selection results exhibit a significant scale bias towards large objects and redundancy in background and symmetrical queries. This results in unsatisfactory performance due to indiscriminative queries. So, what causes these issues and how can we mitigate them?

We attribute these issues to two types of redundancy in the detection transformer: \textbf{encoding redundancy} and \textbf{selection redundancy}. It is generally agreed that image foreground contributes more discriminative features for determining object categories and locations than background \cite{li2023lite, roh2021sparse, zheng2023less}.
Therefore, performing self-attention on background queries may introduce irrelevant and indiscriminative information, which leads to the encoding redundancy. \Cref{tab:pre-ablation of dino query number on mssd} shows that DETR-like methods can still benefit from more two-stage queries, even though the number of them has been much larger than that of actual objects. This indicates that the queries selected for two-stage initialization do not exactly match one-to-one with actual objects, \ie selection redundancy. These two redundancies result in a heavy computational burden as well as indiscriminative queries.
\begin{table}
  \centering
  \small
  \setlength\tabcolsep{4.5pt}
  \caption{Pre-ablation studies on query number of DINO on MSSD}
  \label{tab:pre-ablation of dino query number on mssd}
  \begin{tabular}{c|cccccc}
    \toprule
    Two-stage queries & AP & AP$_{50}$ & AP$_{75}$ & AP$_S$ & AP$_M$ & AP$_{L}$ \\ \midrule
    300 & 49.0 & 78.1 & 49.0 & 19.6 & 28.8 & 43.3 \\
    600 & 49.4 & 79.3 & 48.7 & 18.9 & 46.5 & 43.8 \\
    900 & 51.0 & 80.0 & 52.5 & 20.0 & 47.4 & 44.8 \\
    \bottomrule
  \end{tabular}
\end{table}

Numerous efforts have been made to mitigate redundant calculation and select discriminative queries. For instance, Deformable DETR \cite{zhu2020deformable} reduces the complexity from quadratic to linear and explores multi-scale information usage through deformable attention with sparse reference points. Sparse DETR \cite{roh2021sparse} and Focus DETR \cite{zheng2023less} update only foreground queries for encoding efficiency and achieve comparable precision with much fewer self-attention queries. However, existing query filtering methods apply coarse-grained filtering directly to all tokens, disregarding the multi-scale characteristics where high-level tokens embed more abstract semantics while requiring lower computation compared to low-level tokens \cite{li2023lite,hou2023canet}. Moreover, scale independence is essential when evaluating query importance for unbiased query selection, while the above methods select queries based on the foreground confidence, which may favor large-scale objects and result in a semantic imbalance. Consequently, query filtering becomes ambiguous and misleading.

To tackle these challenges, this paper proposes a novel detector with hierarchical salience filtering refinement, named Salience DETR. We introduce a salience-guided supervision that is scale-independent to overcome the scale bias during query filtering. With the proposed supervision, a hierarchical query filtering mechanism is proposed to mitigate encoding redundancy by encoding only selected queries. In order to compensate for the semantic misalignment among queries, we propose three elaborate modules to refine queries from the perspectives of multi-scale features, foreground-background differences and selection strategies.
Extensive experiments confirm the superior performance and minimal computational cost of Salience DETR.

\section{Related Work}
\subsection{End-to-End detection transformer}
DETR (DEtection TRansformer) proposed by Carion \etal \cite{carion2020end} treats detection as a set prediction task and supervises the training with one-to-one matching through the Hungarian algorithm. Various works have been exploring the transformer-based detectors by accelerating training convergence and improving detection performance \cite{zhu2020deformable, liu2021dab,zong2023detrs,ye2023cascade,li2022dn,zhang2022dino}. Deformable DETR \cite{zhu2020deformable} introduces a framework consisting of two-stage pipeline, deformable attention, and multi-scale tokens that is instructive for subsequent DETR-like methods. Condition-DETR \cite{meng2021conditional} focuses on the extremities of objects through conditional spatial queries to address the slow convergence issue. Since the queries in DETR have no explicit physical meanings, Anchor DETR \cite{wang2022anchor} reintroduces the concepts of anchor query to guide the transformer to focus on specific region modes. 
DINO \cite{zhang2022dino} integrates dynamic anchor \cite{liu2021dab} and contrast denoising training \cite{li2022dn} to construct a mainstream detection framework and realizes the first state-of-the-art performance among DETR-like methods on COCO. Several recent works focus on improving performance by incorporating training-only designs, such as group queries \cite{chen2023group}, hybrid query matching \cite{jia2023detrs} and IoU aware BCE loss \cite{cai2023align}, while leaving the inference process unchanged. Most recently, $\mathcal{C}$o-DETR \cite{zong2023detrs} introduces a versatile label assignment manner that adds parallel auxiliary heads during training and achieves state-of-the-art performance on COCO \cite{lin2014microsoft}.
\subsection{Lightweight Detection Transformer}
Despite the promising performance of the transformer, its high calculation complexity and memory cost hinder further applications. As a representative of attention lightweightness, deformable attention \cite{zhu2020deformable} attends sparse spatial samplings to reduce the computational and memory complexity. Efficient-DETR \cite{yao2021efficient} optimizes the structure to reduce encoder and decoder layers while maintaining comparable performance. Recent works focus on reducing the number of queries participating in self-attention in the encoder. In particular, Lite DETR \cite{li2023lite} prioritizes high-level feature updates to reduce the number of queries. PnP-DETR \cite{wang2021pnp} compresses entire feature maps by abstracting them into a fine-grained vector and a coarse background vector. Sparse DETR \cite{roh2021sparse} refines only the top-$\rho$\% tokens for all encoder layers based on DAM results. Focus DETR \cite{zheng2023less} further introduces the foreground token selector integrated with a cascade set to allocate attention to more informative tokens. However, most of the above methods directly integrate sparsity designs for all tokens while neglecting their multi-scale characteristics.

By contrast, our work performs fine-grained query filtering in both encoding layers and token levels with scale-independent salience supervision and semantic alignment, to address the encoding and selection redundancy.

\section{Salience DETR}
\begin{figure*}
  \centering
  \includegraphics[width=\textwidth]{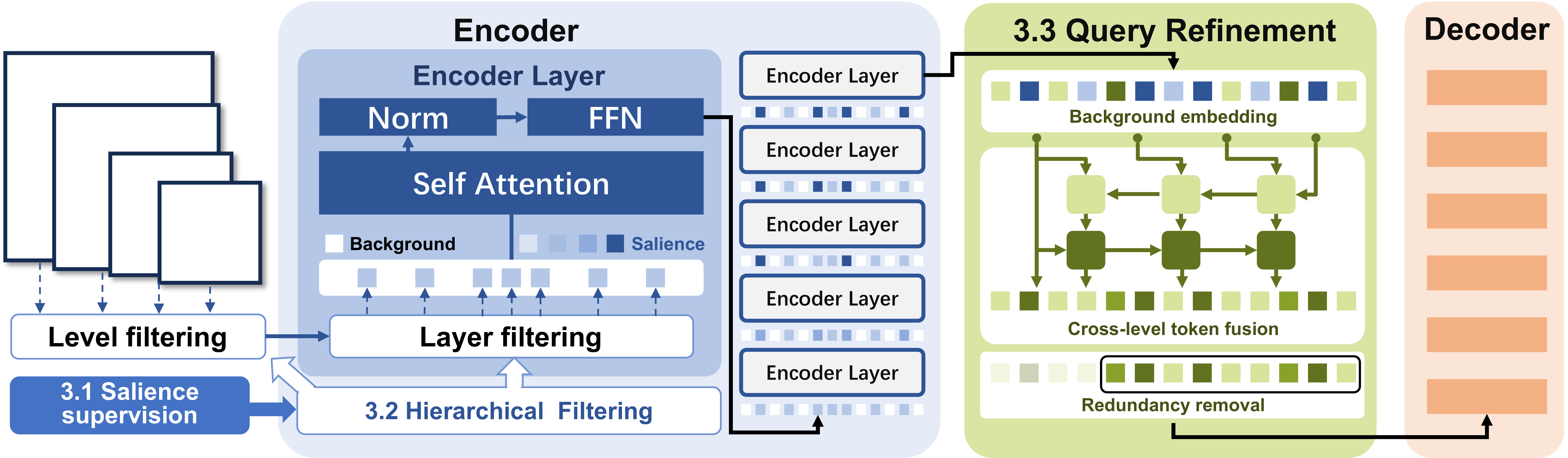}
  \caption{The architecture overview of Salience DETR. We design a hierarchical query filtering for selecting layer-wise and level-wise queries (\Cref{sec:Hierarchical query filtering}) under salience-guided supervision (\Cref{sec:salience-guided supervision}) to mitigate the scale bias in \Cref{fig:query_visualization}. The semantic misalignment among queries is mitigated by query refinement modules (\Cref{sec:query refinement}).}
  \label{fig:salience_detr}
\end{figure*}

As depicted in \Cref{fig:salience_detr}, Salience DETR adopts the high-performance two-stage pipeline. The primary architectural difference between Salience DETR and mainstream two-stage DETR-like methods resides in the transformer encoder and query refinement. Given multi-scale features $\{\boldsymbol f_l\}_{l=1}^L(L=4)$ from the backbone, where $\boldsymbol f_l\in\mathbb R^{C\times H_l\times W_l}$ denotes the feature map downsampled at scale $s_l$, the encoder only updates queries selected by hierarchical query filtering (\Cref{sec:Hierarchical query filtering}) based on salience-guided supervision (\Cref{sec:salience-guided supervision}). The semantic misalignment among queries is mitigated through query refinement modules(\Cref{sec:query refinement}).
\subsection{Salience-guided supervision}
\label{sec:salience-guided supervision}
Query filtering updates the most informative queries to achieve comparable performance with less computational burden, according to the predicted confidence. Drawing inspiration from Focus DETR \cite{zheng2023less}, we provide supervision for the queries at each level in the multi-scale features. Instead of discrete labels $\{0,1\}$ that only classify foreground and background, we construct a scale-independent salience as supervision targets to overcome the scale bias. In particular, each query $t_l^{(i,j)}$ at position $(i,j)$ in the $l$-th feature map corresponds to a coordinate $\boldsymbol c = (x, y)$ in the original image, denoted as $\left(\lfloor \frac{s_l}2\rfloor +i\cdot s_l,\lfloor \frac{s_l}2\rfloor+j\cdot s_l\right)$. The salience confidence $\theta_l^{(i,j)}$ of the query is determined according to the following rules:
\begin{equation}\label{eq:salience confidence}
  \theta_l^{(i,j)}=\left\{\begin{aligned}
    \textcolor{red}{d(\boldsymbol c, \mathcal D_{Bbox})}, &~\boldsymbol c\in\mathcal D_{Bbox} \\
    0~~~~~~~~, &~\boldsymbol c\not\in\mathcal D_{Bbox}
  \end{aligned} \right.
\end{equation}
where $\mathcal D_{Bbox}=(x,y,w,h)$ denotes the ground truth boxes. We highlight the difference between our scale-independent supervision and discrete foreground-background supervision in \textcolor{red}{red}. 
The salience confidence is calculated through the relative distance to object centers:
\begin{equation}
  d(\boldsymbol c, \mathcal D_{Bbox})=1-\sqrt{2\left(\frac{\Delta x}w\right)^2+2\left(\frac{\Delta y}h\right)^2}
\end{equation}
where $\Delta x$ and $\Delta y$ denote the distance between queries and the center of the corresponding bounding boxes.

\begin{figure}
  \centering
  \includegraphics[width=0.45\textwidth]{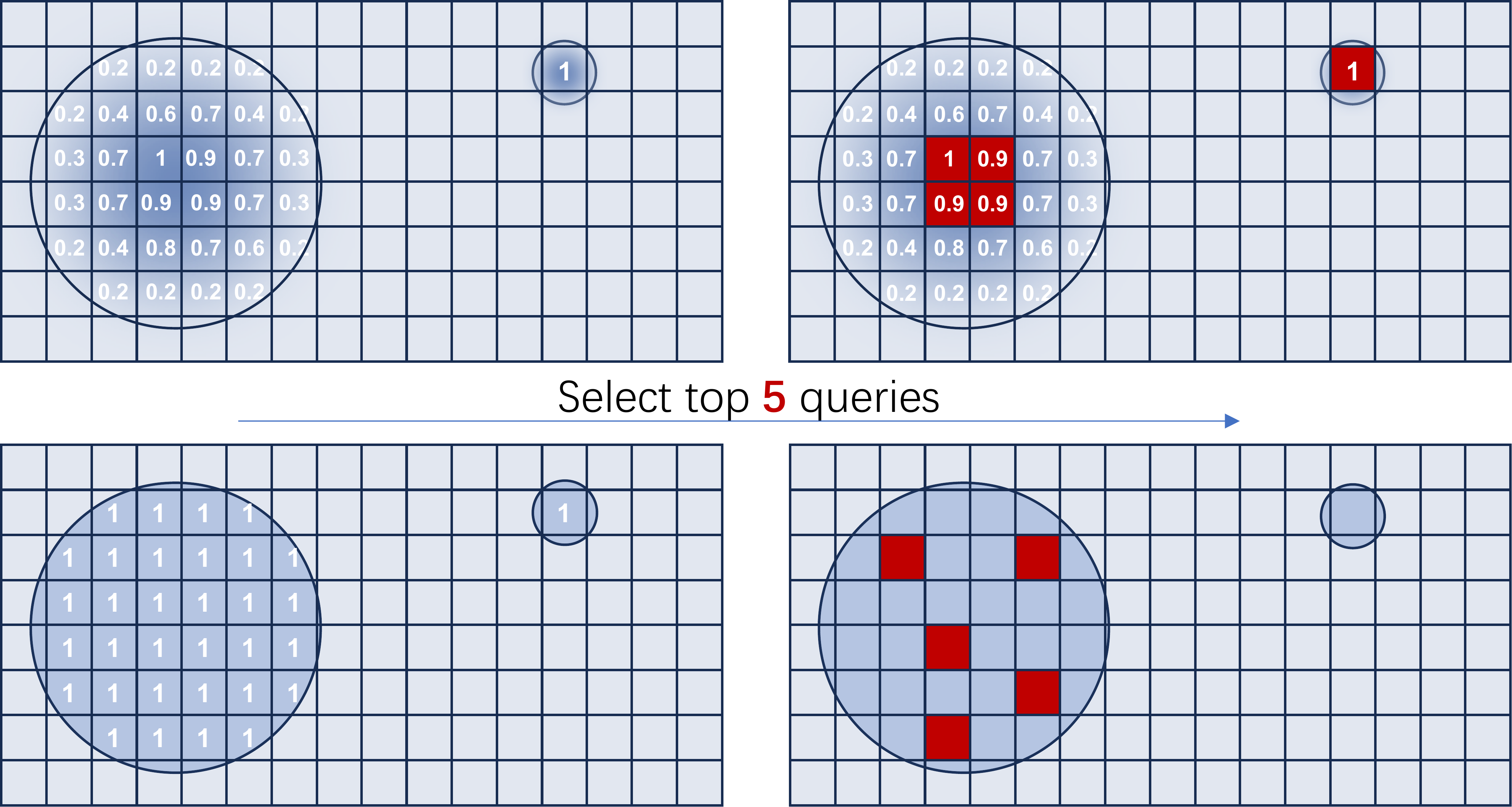}
  \caption{Qualitative illustration of scale-independent supervision (top) and discrete foreground-background supervision (bottom). With salience reducing from the object center to the border, scale-independent supervision balances selected queries even for small-size objects.}
  \label{fig:scale independent supervision}
\end{figure}

\Cref{fig:scale independent supervision} illustrates the comparison between salience supervision and discrete supervision. Rather than indistinguishable labels \cite{roh2021sparse,zheng2023less} that favor large-scale objects (see \Cref{fig:query_visualization}), our salience supervision ensures fine-grained confidence for objects of different scales, bringing more stable filtering results.

\subsection{Heirachical query filtering}
\label{sec:Hierarchical query filtering}
\textbf{Revisting query filtering in Focus DETR.} Focus DETR \cite{zheng2023less} introduces an extra branch that predicts foreground confidence with top-down score modulations on multi-scale features, as follows:
\begin{align}
  \boldsymbol s_{l-1} = \mathbf{MLP}_\mathbf F(\boldsymbol f_{l-1}(1+\mathbf{UP}(\alpha_l*\boldsymbol s_l)))
\end{align}
where $\mathbf{MLP}$ is a global score predictor, $\mathbf{UP}$ is bilinear interpolation, and $\{\alpha_l\}_{l=1}^{L-1}$ are learnable modulation coefficients. Based on this, top $\rho$\% queries with the highest foreground confidence are gathered for transformer encoding and scattered back to update tokens after each encoder layer.

\textbf{Hierarchical query filtering.} Typically, high-level tokens bring less calculation burden while preserving more informative semantics than low-level tokens. Therefore, in addition to the traditional layer-wise filtering, a natural motivation is to introduce level-wise filtering for handling multi-scale characteristics \cite{zheng2023less}. We introduce two sets $\{v_t\}_{t=1}^T$ and $\{w_l\}_{l=1}^L$ as the corresponding filtering ratios, where $T$ and $L$ denote the number of feature levels and encoder layers. For the $t$-th encoder layer and $l$-th feature level, only the top $v_tw_l$ queries are filtered for attention encoding while others are kept unchanged:
\begin{equation}
  q_i=\left\{\begin{aligned}
    \mathrm{Attention}(q_i+pos_i, \boldsymbol q+\boldsymbol{pos}, \boldsymbol q),~\text{if}~q_i\in\Omega_t\\
    q_i~~~~~~~~~~~~~~~~~~~~~,~\text{if}~q_i\not\in\Omega_t
  \end{aligned}\right.
\end{equation}
where $\boldsymbol q=\{q_i\}_{i=1}^{\sum_{l=1}^LH_lW_l}$ and $\boldsymbol{pos}$ are queries and the corresponding position encodings, $\Omega_t$ is the filtered query set in the $t$-th encoder layer.

When using deformable attention \cite{zhu2020deformable} in the encoder, hierarchical query filtering reduces the encoding computation from $O(\sum_{l=1}^LH_lW_lCT(C+KC+5K+3MK))$ to $O(\sum_{1\leqslant l\leqslant L}\sum_{1\leqslant t\leqslant T}w_lv_t H_lW_lC(C+KC+5K+3MK))$, where $M$ and $K$ denote the number of attention head and sampled keys, and the number of queries becomes only $\frac{\|\boldsymbol w\|_1}{T}\frac{\|\boldsymbol v*\boldsymbol s^2\|_1}{\|\boldsymbol s^2\|_1}$ of the original one, with $\boldsymbol s=[s_1,\cdots,s_L]$ denoting the downsample scales of the multi-scale features. 

\subsection{Query refinement}
\label{sec:query refinement}
Due to the differences in the process for selected and unselected queries, the hierarchical query filtering may lead to semantic misalignment among queries. Therefore, we propose three refinement modules (\ie, background embedding, cross-level token fusion, and redundancy removal) to bridge the gap from the perspectives of multi-scale features, foreground-background differences and selection strategies, respectively. 

\textbf{Background embedding.} 
Given self-learned row embeddings and column embeddings $\boldsymbol r,\boldsymbol c \in\mathbb R^{n\times m}$, where $n$ and $m$ denote the number of embeddings and embedding dimensions respectively, we consider building relative and absolute embeddings to refine queries. The former encodes token $\boldsymbol f_l^{(i,j)}$ with relative indexed elements $\boldsymbol r^{(in/H_l)}$ and $\boldsymbol c^{(jn/W_l)}$ thorugh interpolation.
\begin{equation}
  \boldsymbol b_l=\mathop{\mathrm{Interp}}\limits_{(n,n)\rightarrow (H_l, W_l)} (\boldsymbol r\otimes \boldsymbol c)
\end{equation}
where $\otimes$ denotes outer product, $\boldsymbol b_l\in\mathbb R^{C\times H_l\times W_l}$. The latter directly encodes $\boldsymbol f_l^{(i,j)}$ by concatenating $\boldsymbol r^{(i)}$ and $\boldsymbol c^{(j)}$.
\begin{align}
  \boldsymbol b_l^{(i,j)}=\mathrm{Concat}(\boldsymbol r^{(i)}, \boldsymbol c^{(j)})
\end{align}
Then the embedding is added to unselected queries for refinement. In our experiments (see \Cref{tab:comparison of embedding strategies on mssd}), the absolute embedding achieves a higher detection performance and we choose it as the background embedding of Salience DETR. 

\begin{figure}
  \centering
  \includegraphics[width=0.45\textwidth]{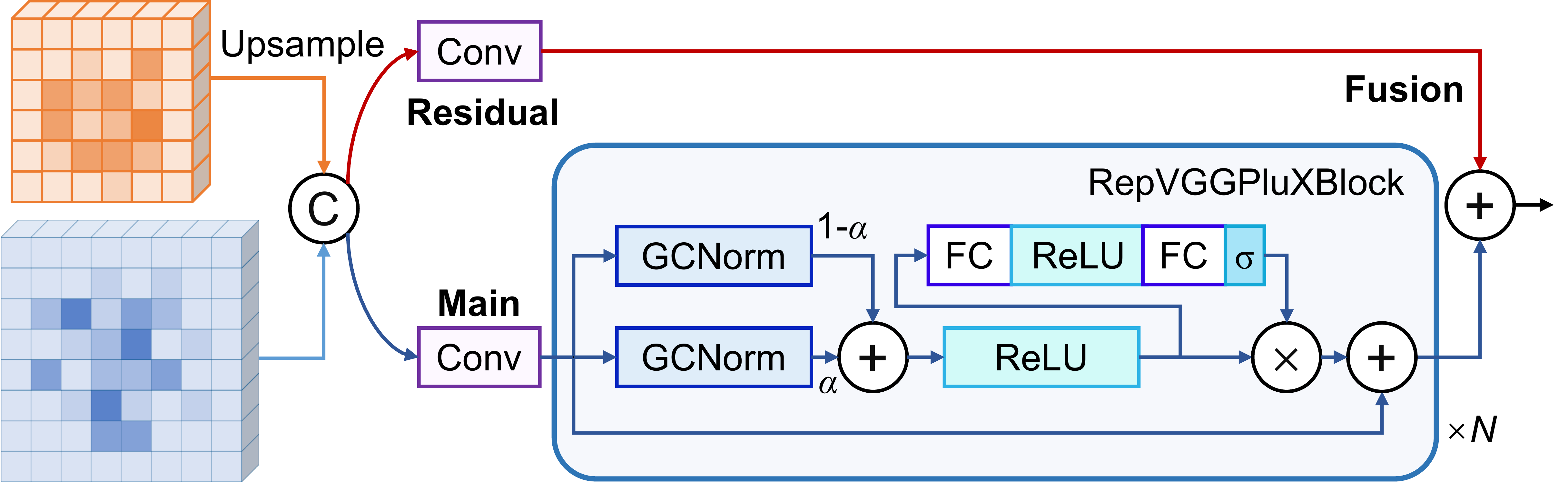}
  \caption{Cross-level token fusion}
  \label{fig:cross-level token fusion}
\end{figure}

\textbf{Cross-level token fusion.} Towards the semantic misalignment of queries at different levels due to the level-specific filtering ratios, we propose a token fusion module that leverages a path aggregation structure \cite{liu2018path} to handle cross-level information interaction. In the module, adjacent tokens are fused through a proposed RepVGGPluXBlock, as shown in \Cref{fig:cross-level token fusion}. For adjacent tokens $\boldsymbol f_l$ and $\boldsymbol f_h$, the calculation is formulated as:
\begin{align}
  \boldsymbol f_I^{(0)}&=\mathbf{Conv}(\mathbf{Concat}(\boldsymbol f_l, \mathbf{UP}(\boldsymbol f_h)))\\
  \boldsymbol f_M^{(n)}&=\mathcal R(\alpha\mathbf{GC}(\boldsymbol f_I^{(n)})+(1-\alpha)\mathbf{GC}(\boldsymbol f_I^{(n)}))\\
  \boldsymbol f_I^{(n+1)}&=\boldsymbol f_M^{(n)}\otimes(\sigma(\mathbf{FC}(\mathcal R(\mathbf{FC}(\boldsymbol f_M^{(n)})))))+\boldsymbol f_I^{(n)}
\end{align}
where $\mathbf{GC}$, $\mathbf{FC}$, $\mathbf{Conv}$, $\mathcal R$, $\sigma$ denote group convolution with batch normalization, dense connection, convolution, ReLU and sigmoid function, respectively. The final refined tokens $\boldsymbol f_I^{(N)}$ in the main branch are added with input tokens through a residual branch, as follows. 
\begin{equation}
  \boldsymbol f_O=\boldsymbol f_I^{(N)}+\mathbf{Conv}(\mathbf{Concat}(\boldsymbol f_l, \mathbf{UP}(\boldsymbol f_h)))
\end{equation}
where $N$ denotes the number of RepVGGPluXBlock.

\textbf{Redundancy removal for two-stage queries.} For those similar objects, especially small-sized objects, the two-stage selection strategy in DETR tends to keep many redundant queries due to their poor discrimination. In addition, one-to-one matching only provides regression supervision on a few positive queries, while massive unsupervised background queries often distribute uniformly like a grid (see \Cref{fig:query_visualization}). Therefore, the transformer decoder suffers from a poor two-stage initialization.
Here we simply remove redundancy through non-maximum suppression, and we expect end-to-end solutions to be proposed to deal with the issue. Specifically, we construct a bounding box $\boldsymbol b_l^{(i,j)}$ with a distance from the center to the border set to 1 for each selected queries. Then, NMS \cite{ren2015faster} is applied to the bounding boxes in both image-wise and level-wise manners.
\begin{equation}
  Bbox_{l}^{(i,j)}=[i-1,j-1,i+1,j+1]
\end{equation}

\subsection{Optimization}
\label{sec:optimization}
Similar to other DETR-like detectors, our model is trained with a multi-task loss function, defined as follows:
\begin{align}
  \mathcal L_{total}&=\lambda_m\mathcal L_m+\lambda_{dn}\mathcal L_{dn}+\lambda_{enc}\mathcal L_{enc}+\lambda_f\mathcal L_f\\
  \mathcal L_f&=-\alpha_f(1-p_f)^\gamma\log(p_f)
\end{align}
where $\mathcal L_f$ denotes the scale-independent supervision loss function, $\alpha_f=0.25$ and $\gamma=2$ are focal parameters, and $p_{f}=\hat\theta\theta+(1-\hat\theta)(1-\theta)$, where $\theta_l^{(i,j)}$ is our proposed salience confidence in \eqref{eq:salience confidence}.

\section{Experiments and Discussions}
\label{sec:experiments and discussions}
This section demonstrates that Salience DETR achieves comparable performance with fewer FLOPs on task-specific and generic detection tasks through quantitative and qualitative analysis, and evaluates the effectiveness of the proposed components through ablation studies.

\subsection{Experimental Setup}
\label{sec:experimental setup}
\textbf{Dataset and Evaluation Metrics.} The evaluation datasets include three task-specific detection datasets (ESD \cite{hou2023canet}, CSD \cite{wang2023surface}, our self-built mobile screen surface dataset (MSSD)), and benchmark COCO 2017 \cite{lin2014microsoft}. Details of them are listed in \Cref{tab:statistics of the three task-specific evaluation datasets}.
\begin{table}[h]
  \small
  \centering
  \setlength\tabcolsep{3pt}
  \caption{Statistics of the three task-specific evaluation datasets}
  \label{tab:statistics of the three task-specific evaluation datasets}
  \begin{tabular}{c|cccccc}
  \toprule
  Dataset & \#ann & \#class & \#train & \#test & Resolution & Obj/img \\ \midrule
  CSD & 4983 & 3 & 373 & 94 & 1024$\times$1024 & 10.67\\
  ESD & 6075 & 2 & 448 & 49 & 3620$\times$3700 & 12.22\\
  MSSD & 93343 & 4 & 962 & 106 & 5120$\times$5120 & 87.40\\ 
  COCO & 896782 & 80 & 118287 & 5000 & - & 7.27 \\
  \bottomrule
  \end{tabular}
\end{table}
We evaluate performance using the standard average precision (AP) and average recall (AR) metrics \cite{lin2014microsoft}.

\textbf{Implementation Details.} The implementation details of Salience DETR align with other DETR-like detectors. We train our model with NVIDIA RTX 3090 GPU (24GB) using the AdamW \cite{kinga2015method} optimizer with a weight decay of $1\times 10^{-4}$. The initial learning rate is set to $1\times 10^{-5}$ for the backbone and $1\times 10^{-4}$ for other parts, which decreases at later stages by 0.1. The batch size per GPU is set to 2. Considering dataset scales, the training epochs on CSD, ESD, and MSSD are 60, 60, 120, respectively. For COCO 2017, we report the results of 12 epochs. The loss coefficients of the salience supervision $\lambda_f$ is set to 2. Other methods are evaluated on detrex \cite{ren2023detrex} and MMDetection \cite{mmdetection} toolbox. 
\subsection{Comparison with State-of-the-art Methods}
\label{sec:comparison with state-of-the-art methods}
\textbf{Comparison on ESD.}
\Cref{tab:quantitative comparison on esd} shows the quantitative comparison on ESD between our Salience DETR and other detectors \cite{ren2015faster,zhu2020autoassign,ye2023cascade,chen2019hybrid,lin2017focal,ge2021yolox,zhu2020deformable,liu2021dab,li2022dn,zhang2022dino,jia2023detrs,zheng2023less}. It can be seen that Salience DETR outperforms the comparison methods under most standard metrics. Notably, considering a strict IoU threshold of 75\%, Salience DETR suppresses the second best method with a large margin of 1.8\% and becomes the only method with AP$_{75}$ over 40\%.
\begin{table*}[htbp]
  \setlength\tabcolsep{4.35pt}
  \small
  \centering
  \caption{Quantitative comparison on ESD. The first and second best results are marked in \textcolor{red}{\textbf{Red}} and \textcolor{blue}{\textbf{Blue}}, respectively.}
  \label{tab:quantitative comparison on esd}
  \begin{tabular}{cc|cc|cccccccccc}
  \toprule
  Methods & Pub'Year & Backbone & Epochs & AP & AP$_{50}$ & AP$_{75}$ & AP$_S$ & AP$_M$ & AP$_L$ & AR & AR$_S$ & AR$_M$ & AR$_L$ \\ \midrule
  Faster RCNN \cite{ren2015faster,he2016deep} & NIPS'2015 & ResNet50 & 60 & 43.5 & 84.4 & \textcolor{blue}{\textbf{39.6}} & 12.5 & 44.2 & 52.5 & 52.0 & 13.1 & 53.2 & 59.0 \\
  AutoAssign \cite{zhu2020autoassign} & Arxiv'2020 & ResNet50 & 60 & 44.2 & 86.0 & 37.5 & 15.4 & 44.8 & 51.8 & 55.7 & 24.8 & 56.2 & 62.5 \\
  Cascade RCNN \cite{ye2023cascade} & CVPR'2018 & ResNet50 & 60 & 44.5 & 85.8 & 39.0 & 14.8 & 44.6 & 55.6 & 54.4 & 17.3 & 55.0 & 62.3 \\
  HTC \cite{chen2019hybrid} & CVPR'2019 & ResNet50 & 60 & \textcolor{blue}{\textbf{45.4}} & 87.1 & 37.3 & 15.3 & \textcolor{blue}{\textbf{45.7}} & 54.7 & 55.4 & 26.3 & 55.7 & 62.3 \\
  RetinaNet \cite{lin2017focal,tan2019efficientnet} & ICML'2019 & EfficientNet & 60 & 43.4 & 87.2 & 35.7 & 13.4 & 43.2 & 53.2 & 53.4 & 23.8 & 53.6 & 63.4 \\
  YOLOX \cite{ge2021yolox} & Arxiv'2021 & CSPDarknet & 300 & 41.6 & 79.6 & 36.9 & 14.9 & 43.4 & 50.0 & 54.3 & \textcolor{red}{\textbf{36.9}} & 54.2 & 60.8 \\ \midrule
  Def-DETR \cite{zhu2020deformable} & ICLR'2020 & ResNet50 & 300 & 42.3 & 85.8 & 33.8 & \textcolor{red}{\textbf{15.5}} & 42.4 & 51.2 & 53.6 & 24.2 & 53.9 & 61.8 \\
  DAB-Def-DETR \cite{liu2022dab} & ICLR'2021 & ResNet50 & 90 & 39.8 & 84.6 & 30.4 & 7.4 & 40.0 & 55.5 & 52.4 & 9.4 & 52.9 & 66.6 \\
  DN-Def-DETR \cite{li2022dn} & CVPR'2022 & ResNet50 & 60 & 40.9 & 86.2 & 32.9 & 10.5 & 40.9 & 54.2 & 56.1 & 18.1 & 56.5 & 65.9 \\
  DINO \cite{zhang2022dino} & ICLR'2022 & ResNet50 & 60 & 42.5 & \textcolor{blue}{\textbf{87.8}} & 34.1 & 12.3 & 42.1 & \textcolor{blue}{\textbf{57.5}} & 55.0 & 19.4 & 54.8 & 67.2 \\
  H-Def-DETR \cite{jia2023detrs} & CVPR'2023 & ResNet50 & 60 & 41.9 & 87.7 & 33.3 & 12.0 & 41.8 & 55.3 & 54.6 & 28.1 & 54.0 & 66.6 \\
  Focus-DETR \cite{zheng2023less} & ICCV'2023 & ResNet50 & 60 & 43.3 & \textcolor{red}{\textbf{88.6}} & 34.3 & 11.7 & 43.3 & \textcolor{blue}{\textbf{57.5}} & \textcolor{blue}{\textbf{57.2}} & 21.9 & \textcolor{blue}{\textbf{57.0}} & \textcolor{blue}{\textbf{67.4}} \\
  Salience DETR & ours & ResNet50 & 60 & \textcolor{red}{\textbf{46.5}} & \textcolor{red}{\textbf{88.6}} & \textcolor{red}{\textbf{41.4}} & \textcolor{blue}{\textbf{15.4}} & \textcolor{red}{\textbf{46.8}} & \textcolor{red}{\textbf{57.7}} & \textcolor{red}{\textbf{58.4}} & \textcolor{blue}{\textbf{34.0}} & \textcolor{red}{\textbf{58.2}} & \textcolor{red}{\textbf{69.5}} \\ 
  \bottomrule
  \end{tabular}
\end{table*}

\textbf{Comparison on CSD.}
\Cref{tab:quantitative comparison on csd} shows the results on CSD. It is worth noting that CSD is challenging for DETR-like methods since it lacks large-size objects suitable for DETR detection. Comprehensively speaking, our Salience DETR achieves the highest performance considering average precision (+0.2\%) and average recall (+0.2\%) and outperforms the latest DETR-like methods \cite{zhang2022dino,zheng2023less,jia2023detrs}.
\begin{table*}[htbp]
  \small
  \setlength\tabcolsep{4.35pt}
  \centering
  \caption{Quantitative comparison on CSD. The first and second best results are marked in \textcolor{red}{\textbf{Red}} and \textcolor{blue}{\textbf{Blue}}, respectively.}
  \label{tab:quantitative comparison on csd}
  \begin{tabular}{cc|cc|cccccccccc}
  \toprule
  Methods & Pub'Year & Backbone & Epochs & AP & AP$_{50}$ & AP$_{75}$ & AP$_S$ & AP$_M$ & AP$_L$ & AR & AR$_S$ & AR$_M$ & AR$_L$ \\ \midrule
  Faster RCNN \cite{ren2015faster,he2016deep} & NIPS'2015 & ResNet50 & 60 & 52.4 & \textcolor{red}{\textbf{93.2}} & 51.9 & 48.5 & 39.7 & \textcolor{red}{\textbf{70.0}} & 59.9 & 57.4 & 44.0 & \textcolor{red}{\textbf{70.0}} \\
  AutoAssign \cite{zhu2020autoassign} & Arxiv'2020 & ResNet50 & 60 & 51.1 & 92.5 & 49.5 & 47.2 & 40.1 & \textcolor{blue}{\textbf{60.0}} & 60.1 & 56.7 & 58.1 & 60.0 \\
  RetinaNet \cite{lin2017focal,tan2019efficientnet} & ICML'2019 & EfficientNet & 60 & 50.5 & 90.4 & 54.1 & 44.0 & \textcolor{blue}{\textbf{40.9}} & 5.0 & 59.8 & 54.3 & 58.6 & 20.0 \\
  YOLOX \cite{ge2021yolox} & Arxiv'2021 & CSPDarknet & 300 & 47.5 & 90.0 & 42.4 & 47.0 & 36.5 & 0 & 56.1 & 56.6 & 40.6 & 0 \\
  TOOD \cite{feng2021tood} & ICCV'2021 & ResNet50 & 60 & 52.9 & \textcolor{blue}{\textbf{92.9}} & 52.7 & 47.7 & \textcolor{red}{\textbf{41.8}} & 14.5 & 60.6 & 57.0 & 45.6 & 60.0 \\ \midrule
  Def-DETR \cite{zhu2020deformable} & ICLR'2020 & ResNet50 & 300 & 43.7 & 86.2 & 36.6 & 40.5 & 34.9 & 10.0 & 56.0 & 55.3 & 62.1 & 10.0 \\
  DAB-Def-DETR \cite{liu2022dab} & ICLR'2021 & ResNet50 & 90 & 52.9 & 91.2 & 55.0 & 50.3 & 39.4 & 0 & 62.5 & 60.6 & 58.1 & 0 \\
  DN-Def-DETR \cite{li2022dn} & CVPR'2022 & ResNet50 & 60 & 49.9 & 88.0 & 51.2 & 47.6 & 37.7 & 0 & 63.7 & 61.0 & \textcolor{red}{\textbf{76.3}} & 0 \\
  DINO \cite{zhang2022dino} & ICLR'2022 & ResNet50 & 60 & \textcolor{blue}{\textbf{53.0}} & 90.8 & 55.5 & 50.9 & 39.6 & 0 & 64.0 & 63.1 & 59.1 & 0 \\
  H-Def-DETR \cite{jia2023detrs} & CVPR'2023 & ResNet50 & 60 & \textcolor{blue}{\textbf{53.0}} & 90.6 & \textcolor{blue}{\textbf{55.7}} & \textcolor{red}{\textbf{51.2}} & 39.2 & 6.7 & 63.2 & 62.2 & 45.3 & 30.0 \\
  Focus-DETR \cite{zheng2023less} & ICCV'2023 & ResNet50 & 60 & 52.3 & 91.2 & \textcolor{red}{\textbf{55.9}} & 50.3 & 39.2 & 0.9 & \textcolor{blue}{\textbf{65.3}} & \textcolor{blue}{\textbf{64.1}} & 71.2 & 60.0 \\
  Salience DETR & ours & ResNet50 & 60 & \textcolor{red}{\textbf{53.2}} & 92.5 & 55.1 & \textcolor{blue}{\textbf{51.0}} & \textcolor{blue}{\textbf{40.9}} & 0 & \textcolor{red}{\textbf{66.5}} & \textcolor{red}{\textbf{65.7}} & \textcolor{blue}{\textbf{74.3}} & 0 \\ \bottomrule
  \end{tabular}
\end{table*}

\textbf{Comparison on MSSD.}
The MSSD dataset, collected from industrial production lines by ourselves, contains massive small-sized and weak objects with low contrast and indistinguishable contours. Therefore, discriminative query selection is important for detecting this dataset. As can be seen from \Cref{tab:quantitative comparison on mssd}, with hierarchical salience filtering refinement, our Salience DETR achieves a superior AP$_{75}$ of 61.9\% with a large margin of 9.4\% compared to the second best result and improves AP by 4.4\%. Moreover, due to the salience-guided supervision that is scale-independent, queries in Salience DETR match small-sized objects better and thus our Salience DETR improves AP$_S$ and AR$_S$ by 8.7\% and 9.1\% compared to DINO, respectively.
\begin{table*}[htbp]
  \small
  \setlength\tabcolsep{4.35pt}
  \centering
  \caption{Quantitative comparison on MSSD. The first and second best results are marked in \textcolor{red}{\textbf{Red}} and \textcolor{blue}{\textbf{Blue}}, respectively.}
  \label{tab:quantitative comparison on mssd}
  \begin{tabular}{cc|cc|cccccccccc}
  \toprule
  Methods & Pub'Year & Backbone & Epochs & AP & AP$_{50}$ & AP$_{75}$ & AP$_S$ & AP$_M$ & AP$_L$ & AR & AR$_S$ & AR$_M$ & AR$_L$ \\ \midrule
  Faster RCNN \cite{ren2015faster,he2016deep} & NIPS'2015 & ResNet50 & 120 & 44.5 & 65.3 & 48.6 & 18.0 & 36.7 & 40.3 & 51.1 & 21.3 & 44.5 & 46.3 \\
  AutoAssign \cite{zhu2020autoassign} & Arxiv'2020 & ResNet50 & 120 & 38.4 & 56.2 & 40.5 & 6.3 & 21.6 & 40.9 & 45.7 & 9.9 & 37.0 & 46.6 \\
  Cascade RCNN \cite{ye2023cascade} & CVPR'2018 & ResNet50 & 120 & 47.5 & 69.5 & 52.1 & \textcolor{blue}{\textbf{23.5}} & 42.0 & 42.0 & 54.1 & \textcolor{blue}{\textbf{26.1}} & 51.3 & 48.3 \\
  YOLOX \cite{ge2021yolox} & Arxiv'2021 & CSPDarknet & 500 & 41.4 & 67.5 & 39.2 & 13.6 & 35.5 & 38.7 & 53.7 & 21.5 & 46.8 & 48.6 \\
  HTC \cite{chen2019hybrid} & CVPR'2019 & ResNet50 & 120 & 47.5 & 68.1 & 52.5 & 19.9 & 33.2 & 41.6 & 53.7 & 22.7 & 49.6 & 46.0 \\ \midrule
  Def-DETR \cite{zhu2020deformable} & ICLR'2020 & ResNet50 & 300 & 33.0 & 54.3 & 32.0 & 9.8 & 11.1 & 33.3 & 39.8 & 13.6 & 18.9 & 38.5 \\
  DAB-Def-DETR \cite{liu2022dab} & ICLR'2021 & ResNet50 & 120 & 33.7 & 60.0 & 31.0 & 15.9 & 26.7 & 29.0 & 46.2 & 19.9 & 38.9 & 39.8 \\
  DN-Def-DETR \cite{li2022dn} & CVPR'2022 & ResNet50 & 120 & 45.6 & 74.1 & 44.9 & 18.6 & 31.9 & 41.0 & 53.9 & 21.7 & 50.1 & 47.7 \\
  DINO \cite{zhang2022dino} & ICLR'2022 & ResNet50 & 120 & \textcolor{blue}{\textbf{51.0}} & \textcolor{red}{\textbf{80.0}} & \textcolor{blue}{\textbf{52.5}} & 20.0 & \textcolor{blue}{\textbf{47.4}} & \textcolor{red}{\textbf{44.8}} & \textcolor{blue}{\textbf{61.3}} & 23.9 & \textcolor{red}{\textbf{60.1}} & \textcolor{red}{\textbf{52.4}} \\
  H-Def-DETR \cite{jia2023detrs} & CVPR'2023 & ResNet50 & 120 & 46.9 & 76.8 & 47.1 & 20.3 & 45.3 & 40.7 & 57.1 & 25.8 & 57.5 & 49.0 \\
  Focus-DETR \cite{zheng2023less} & ICCV'2023 & ResNet50 & 120 & 49.2 & \textcolor{blue}{\textbf{79.3}} & 47.9 & 18.3 & 40.7 & 43.5 & 59.4 & 22.1 & 53.3 & 51.1 \\
  Salience DETR & ours & ResNet50 & 120 & \textcolor{red}{\textbf{55.4}} & 78.2 & \textcolor{red}{\textbf{61.9}} & \textcolor{red}{\textbf{28.7}} & \textcolor{red}{\textbf{47.5}} & \textcolor{blue}{\textbf{44.5}} & \textcolor{red}{\textbf{65.4}} & \textcolor{red}{\textbf{33.0}} & \textcolor{blue}{\textbf{59.8}} & \textcolor{blue}{\textbf{52.1}} \\ \bottomrule
  \end{tabular}
\end{table*}
\subsection{Ablation Studies}
\label{sec:ablation studies}
Ablation studies are conducted on CSD \cite{wang2023surface} using ResNet50 backbone and \Cref{tab:ablation study results on csd} reports the results. As can be seen, by introducing hierarchical query filtering with scale-independent salience supervision, Salience DETR yields a significant improvement of +1.8 AP. Then the background embedding and the redundancy removal steadily increase AP from 52.0\% to 52.8\% while keeping FLOPs at 168G with no extra computation. Finally, the $6$-th row of \Cref{tab:ablation study results on csd} shows that Salience DETR equipped with the cross-level token fusion achieves +0.4 AP. These results demonstrate the effectiveness of our proposed components.
\begin{table}[htbp]
  \small
  \centering
  \setlength\tabcolsep{1.8pt}
  \caption{Ablation results on CSD. HQF: hierarchical query filtering (\Cref{sec:Hierarchical query filtering}), BE: background embedding, RR: redundancy removal, CTF: cross-level token fusion}
  \label{tab:ablation study results on csd}
  \begin{tabular}{cccc|ccccccc}
  \toprule
  HQF. & BE. & RR. & CTF. & AP & AP$_{50}$ & AP$_{75}$ & AP$_S$ & AP$_M$ & AR & FLOPs \\ \midrule
  \ding{55} & \ding{55} & \ding{55} & \ding{55} & 50.2 & 89.8 & 49.4 & 47.8 & 37.9 & 64.1 & 132 \\
  \checkmark & \ding{55} & \ding{55} & \ding{55} & 52.0 & 89.7 & 54.2 & 49.4 & 39.8 & 65.6 & 168 \\
  \checkmark & \checkmark & \ding{55} & \ding{55} & 52.0 & 90.3 & 49.9 & 50.3 & 39.1 & 64.5 & 168 \\
  \checkmark & \ding{55} & \checkmark & \ding{55} & 52.6 & 90.8 & 52.7 & 50.9 & 39.2 & 64.4 & 168 \\
  \checkmark & \checkmark & \checkmark & \ding{55} & 52.8 & 91.0 & 55.7 & 51.5 & 39.3 & 64.7 & 168 \\
  \checkmark & \checkmark & \checkmark & \checkmark & 53.2 & 92.5 & 55.1 & 51.0 & 40.9 & 66.5 & 201 \\ \bottomrule
  \end{tabular}
\end{table}

\textbf{Effect of scale-independent supervision.}
Unlike assigning queries to feature levels according to their corresponding object scales \cite{zheng2023less}, one of our contributions is introducing a scale-independent supervision that is totally determined by salience. \Cref{tab:quantitative comparison of supervision methods} compares the effect of them, in which we can see that scale-independent supervision brings consistent precision improvements (+1.8\% AP, +0.4\% AP$_{50}$, +1.6\% AP$_{75}$) compared to supervision determined according to object scales, confirming its effectiveness to address scale bias in query filtering and benefit final performance.
\begin{table}[!htbp]
  \centering
  \small
  \setlength\tabcolsep{4.3pt}
  \caption{Quantitative comparison of supervision methods on MSSD. The intervals $\left[[-1, 128], [64, 256], [128, 512], [256, \infty]\right]$ are used in overlap limit range, following Focus-DETR.}
  \label{tab:quantitative comparison of supervision methods}
  \begin{tabular}{c|cccccccccc}
    \toprule
    Methods & AP & AP$_{50}$ & AP$_{75}$ & AP$_S$ & AP$_M$ & AP$_L$ \\ \midrule
    Overlap limit \cite{zheng2023less} & 53.2 & 77.4 & 60.3 & \textbf{29.0} & 39.4 & 43.7 \\ 
    Scale-independent & \textbf{55.0} & \textbf{77.8} & \textbf{61.9} & 28.1 & \textbf{40.7} & \textbf{44.1} \\
    \bottomrule
  \end{tabular}
\end{table}

\textbf{Effect of background embedding.}
We analyze the effect of the proposed two variants of background embeddings that compensate for semantic misalignments in query filtering. As shown in \Cref{tab:comparison of embedding strategies on mssd}, the absolute embedding could boost AP slightly better than the relative embedding. This may be because the query filtering is performed on pixels and absolute embedding provides direct position information and benefits position-related features.
\begin{table}[htbp]
  \centering
  \small
  \setlength\tabcolsep{3.9pt}
  \caption{Comparison of embedding strategies on MSSD}
  \label{tab:comparison of embedding strategies on mssd}
  \begin{tabular}{c|cccccccc}
    \toprule
    Embed Strategies & AP & AP$_{50}$ & AP$_{75}$ & AP$_S$ & AP$_M$ & AP$_L$ \\ \midrule
    Relative embedding & 55.0 & 77.6 & 62.1 & 29.4 & 46.9 & 44.0 \\
    Absolute embedding & 55.2 & 77.6 & 61.3 & 29.6 & 51.4 & 44.1 \\
    \bottomrule
  \end{tabular}
\end{table}

\begin{figure}[htbp]
  \centering
  \subfloat{
    \hspace{-15pt}
    \includegraphics[width=0.25\textwidth]{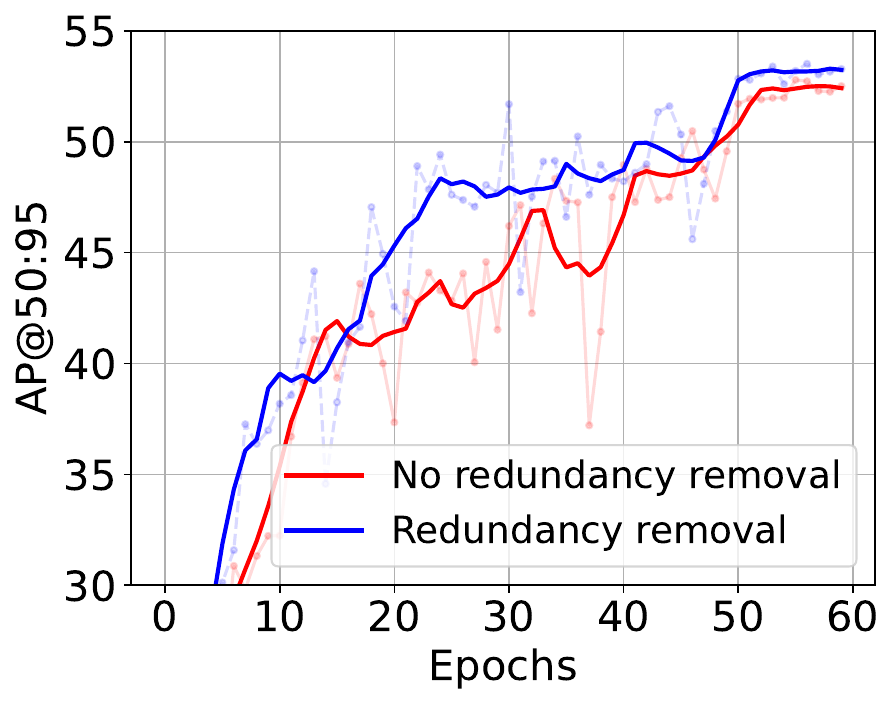}
  }
  \subfloat{
    \hspace{-8pt}
    \includegraphics[width=0.25\textwidth]{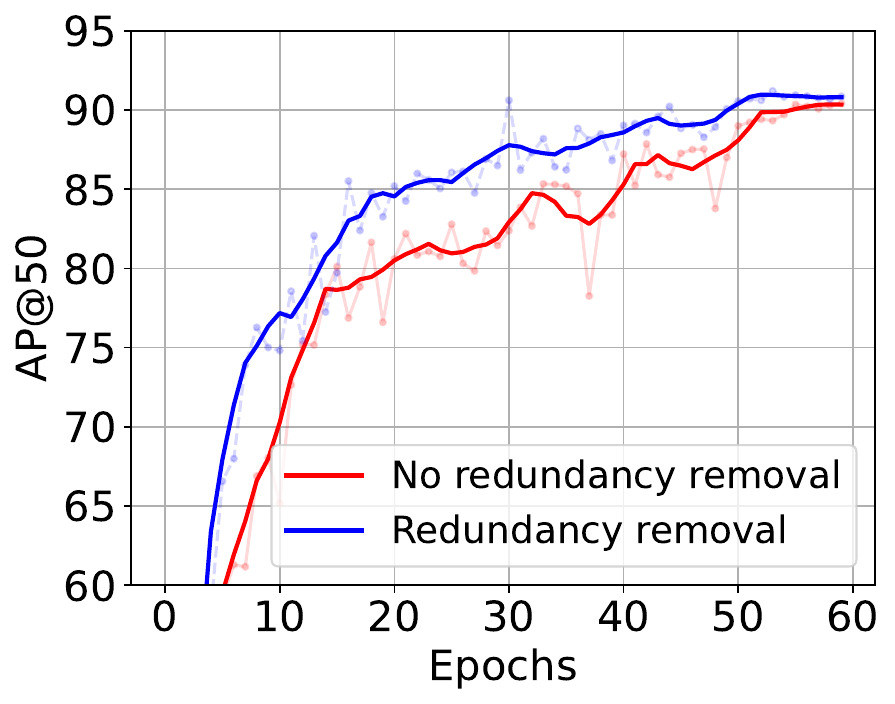}
  }
  \caption{Convergence of Salience DETR}
  \label{fig:convergence of salience detr}
\end{figure}

\textbf{Effect of redundancy removal.}
As mentioned in \Cref{sec:query refinement}, the redundancy removal is critical to stabilizing two-stage initialization. From the evaluation metrics AP and AP$_{50}$ in \Cref{fig:convergence of salience detr}, we can see that the proposed redundancy removal could speed up convergence with a significant margin, especially at early stages. Salience DETR with redundancy removal achieves better performances of +0.7\% AP and +0.4\% AP$_{50}$, which is mainly attributed to the fact that redundancy removal allows the decoder to focus on unique and relevant features provided by more discriminative queries.

\subsection{Scalability of Salience DETR}
\label{sec:scalability of salience-detr}
With designed salience filtering refinement, our Salience DETR can effectively provide well-matched focus for objects in defect detection tasks. Here we illustrate its great scalability to generic large-scale datasets of COCO 2017. As shown in \Cref{tab:quantitative comparison on COCO 2017}, our Salience DETR achieves better performance of 49.2\% AP compared to other methods with less than 70\% FLOPs of DINO \cite{zhang2022dino} under the same setting, demonstrating its superior trade-off between computational complexity and performance.
\begin{table*}[htbp]
  \centering
  \small
  \setlength\tabcolsep{6.2pt}
  \caption{Quantitative comparison on COCO val2017. Since the FLOPs may differ according to the calculation script, we reimplement DINO and report its FLOPs results using the same script with our Salience DETR.}
  \label{tab:quantitative comparison on COCO 2017}
  \begin{tabular}{cc|ccccccccc}
  \toprule
  Method & Pub'Yead & epochs & backbone & AP$\ \uparrow$ & AP$_{50}\uparrow$ & AP$_{75}\uparrow$ & AP$_S\uparrow$ & AP$_M\uparrow$ & AP$_L\uparrow$ & FLOPs$\ \downarrow$\\ \hline
  Conditional-DETR \cite{meng2021conditional} & CVPR'21 & 108 & R50 & 43.0 & 64.0 & 45.7 & 22.7 & 46.7 & 61.5 & -\\
  SAM-DETR \cite{zhang2022accelerating} & CVPR'22 & 50 & R50 & 41.8 & 63.2 & 43.9 & 22.1 & 45.9 & 60.9 & -\\
  Anchor-DETR \cite{wang2022anchor} & AAAI'22 & 50 & R50 & 42.1 & 63.1 & 44.9 & 22.3 & 46.2 & 60.0 & -\\
  Dynamic-DETR \cite{dai2021dynamic} & CVPR'21 & 12 & R50 & 42.9 & 61.0 & 46.3 & 24.6 & 44.9 & 54.4 & -\\
  Sparse-DETR \cite{roh2021sparse} & ICLR'21 & 50 & R50 & 46.3 & 66.0 & 50.1 & 29.0 & 49.5 & 60.8 & -\\
  Efficient-DETR \cite{yao2021efficient} & Arxiv'21 & 36 & R50 & 45.1 & 63.1 & 49.1 & 28.3 & 48.4 & 59.0 & -\\ \midrule
  Def-DETR \cite{zhu2020deformable} & ICLR'20 & 50 & R50 & 46.9 & 65.6 & 51.0 & 29.6 & 50.1 & 61.6 & -\\
  DAB-Def-DETR \cite{liu2021dab} & ICLR'21 & 50 & R50 & 46.8 & 66.0 & 50.4 & 29.1 & 49.8 & 62.3 & -\\
  DN-Def-DETR \cite{li2022dn} & CVPR'22 & 50 & R50 & 48.6 & \textcolor{red}{\textbf{67.4}} & 52.7 & 31.0 & 52.0 & \textcolor{red}{\textbf{63.7}} & -\\
  Focus-DETR \cite{zheng2023less} & CVPR'23 & 12 & R50 & 48.8 & 66.8 & 52.8 & 31.7 & 52.1 & 63.0 & - \\
  H-DETR \cite{jia2023detrs} & CVPR'23 & 12 & R50 & 48.7 & 66.4 & 52.9 & 31.2 & 51.5 & \textcolor{blue}{\textbf{63.5}} & -\\
  DINO \cite{zhang2022dino} & ICLR'22 & 12 & R50 & \textcolor{blue}{\textbf{49.0}} & 66.6 & \textcolor{blue}{\textbf{53.5}} & \textcolor{blue}{\textbf{32.0}} & \textcolor{blue}{\textbf{52.3}} & 63.0 & 291G \\
  Salience DETR & ours & 12 & R50 & \textcolor{red}{\textbf{49.2}} & \textcolor{blue}{\textbf{67.1}} & \textcolor{red}{\textbf{53.8}} & \textcolor{red}{\textbf{32.7}} & \textcolor{red}{\textbf{53.0}} & 63.1 & \textbf{201G} \\ \bottomrule
  \end{tabular}
\end{table*}

\subsection{Visualization}
\label{sec:visualization}
\begin{figure}[htbp]
  \centering
  \hspace{-6pt}
  \subfloat{
    \includegraphics[width=0.105\textwidth]{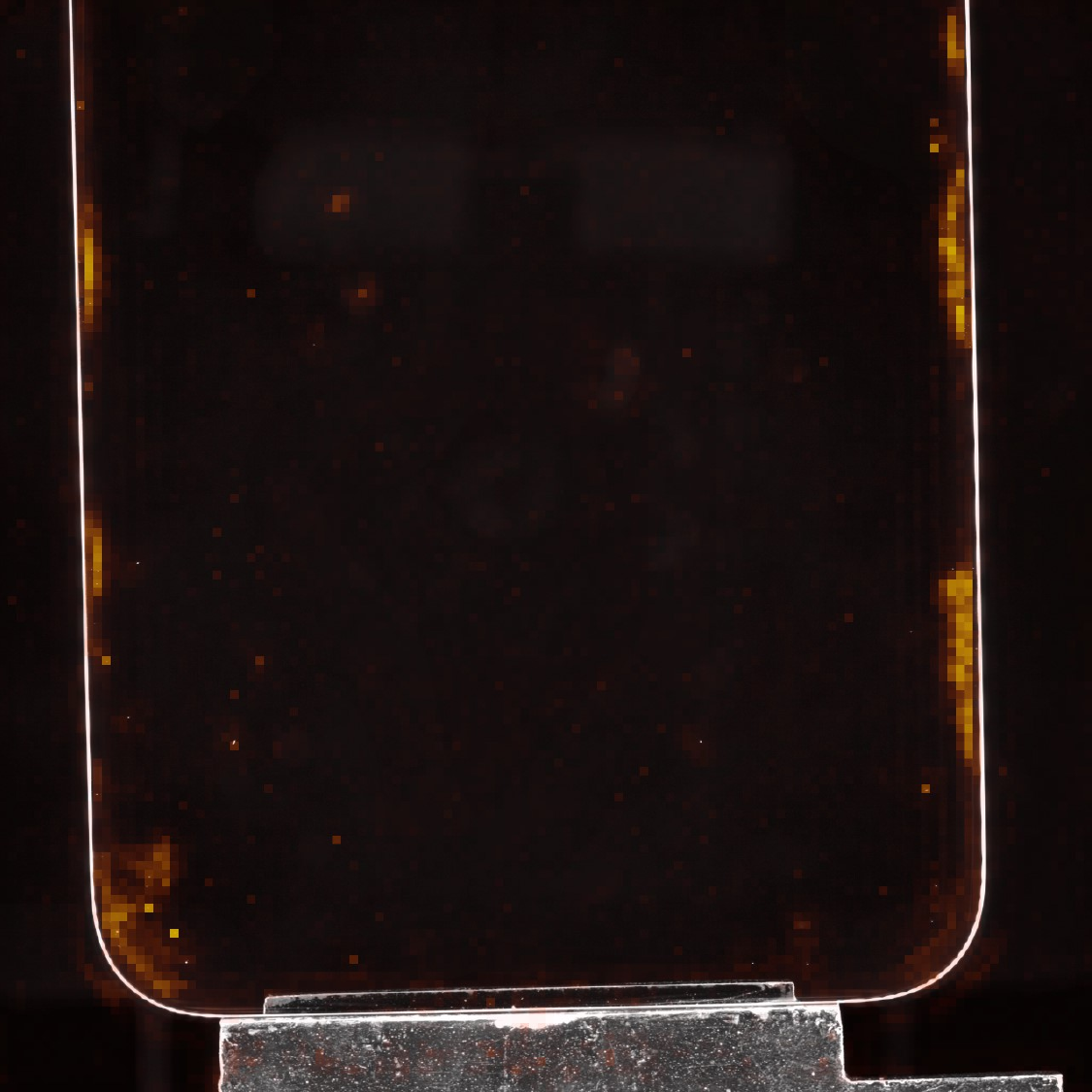}
  }\hspace{-6pt}
  \subfloat{
    \includegraphics[width=0.105\textwidth]{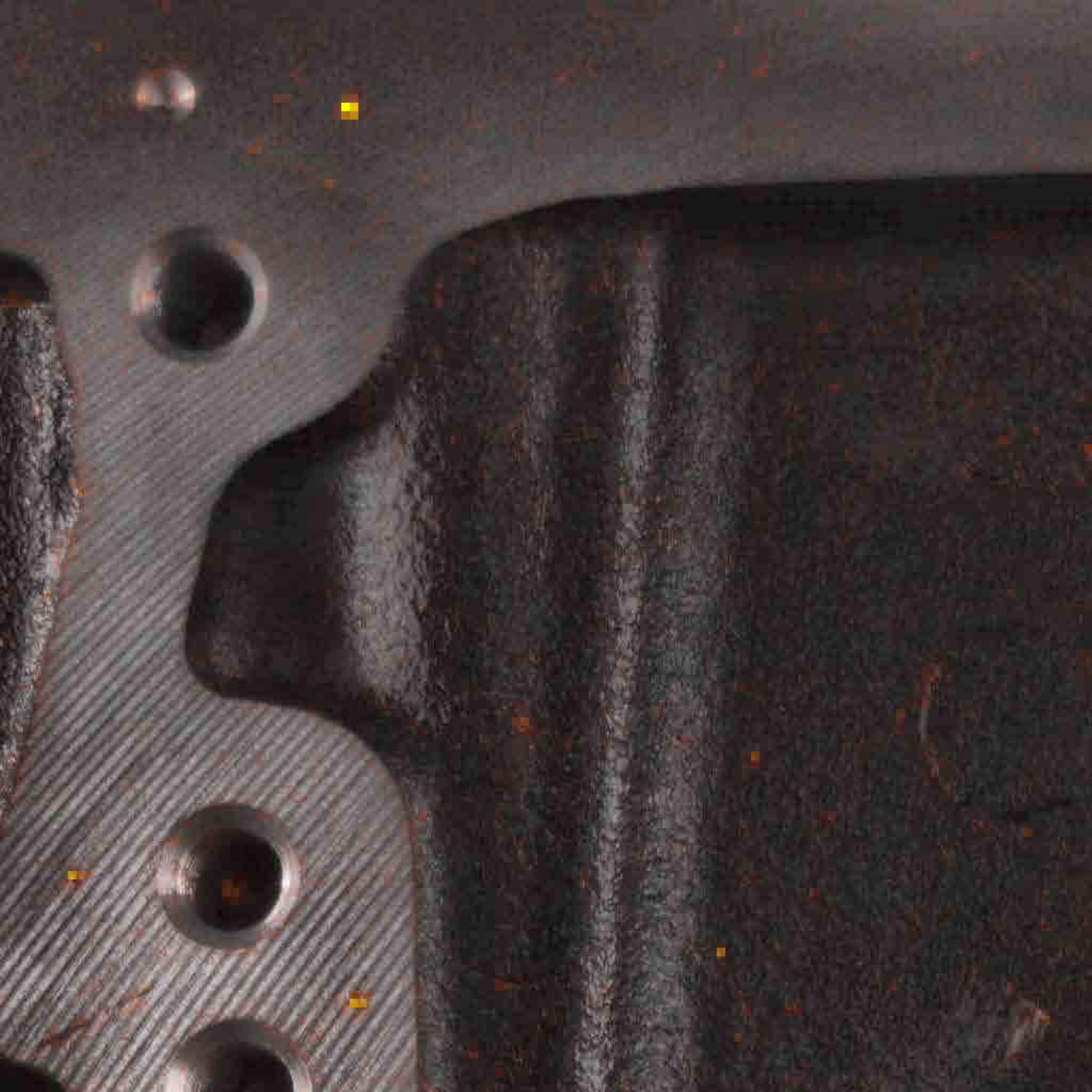}
  }\hspace{-6pt}
  \subfloat{
    \includegraphics[height=0.105\textwidth]{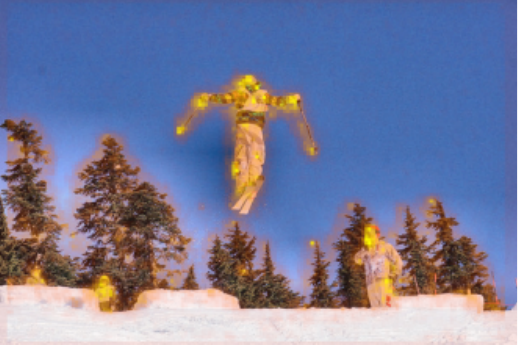}
  }\hspace{-6pt}
  \subfloat{
    \includegraphics[height=0.105\textwidth]{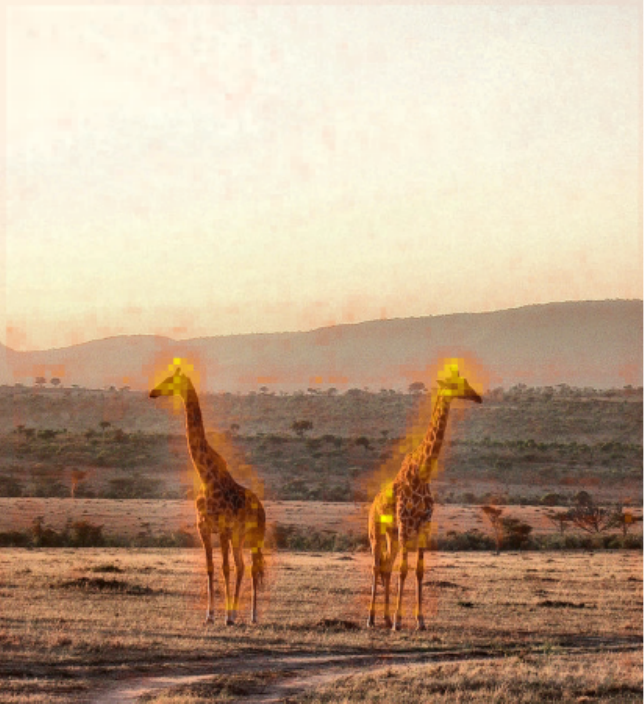}
  }
  \\
  \hspace{-6pt}
  \subfloat{
    \includegraphics[width=0.105\textwidth]{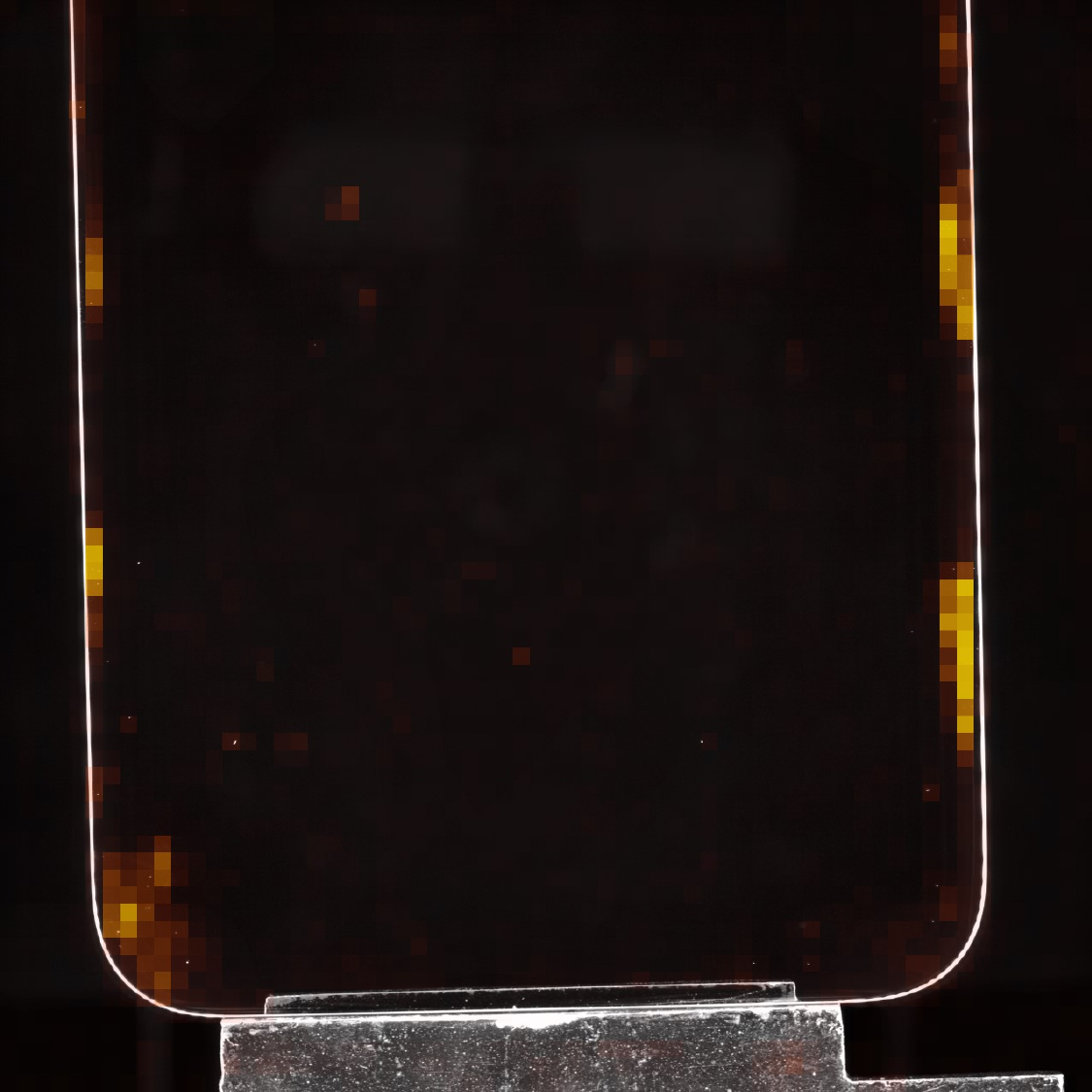}
  }\hspace{-6pt}
  \subfloat{
    \includegraphics[width=0.105\textwidth]{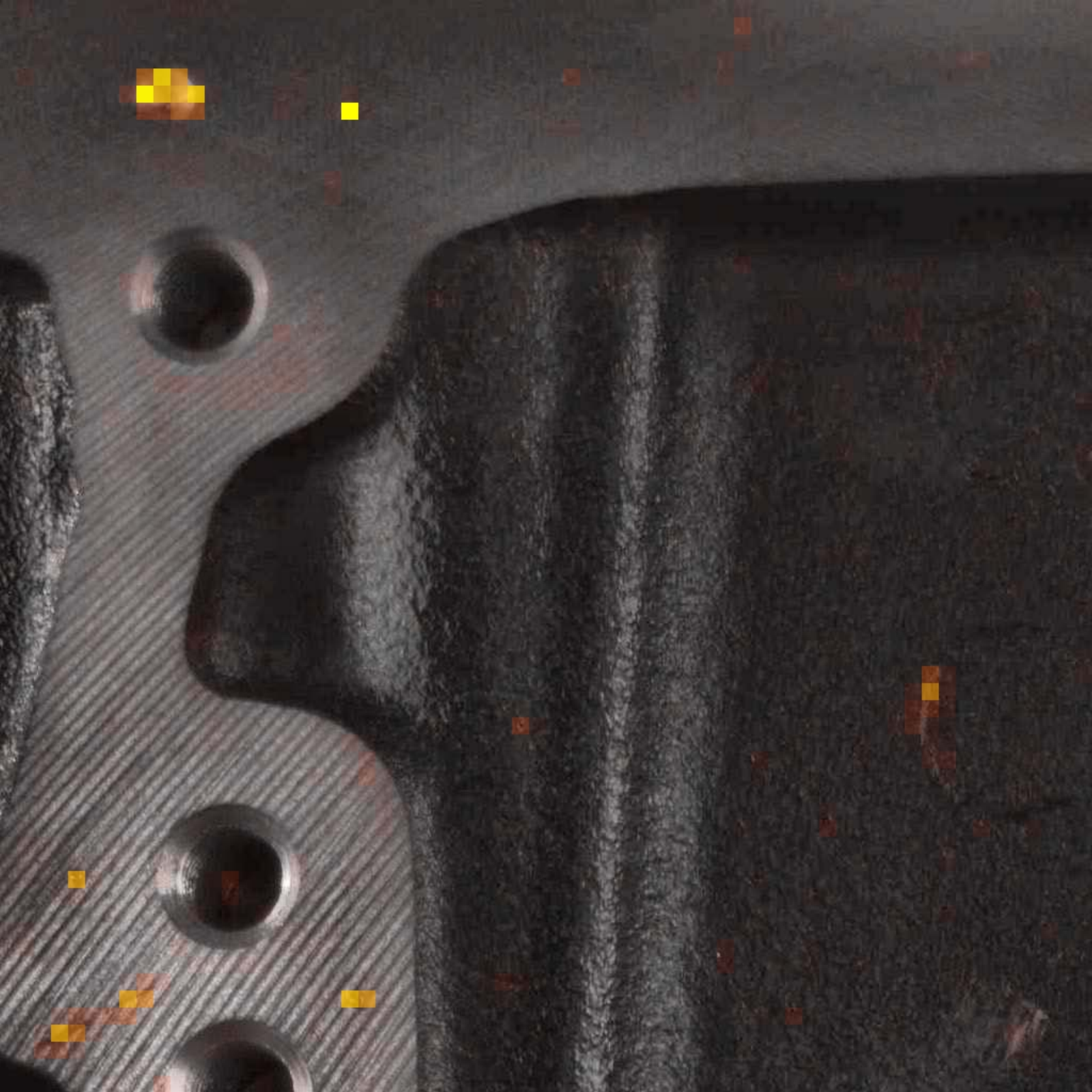}
  }\hspace{-6pt}
  \subfloat{
    \includegraphics[height=0.105\textwidth]{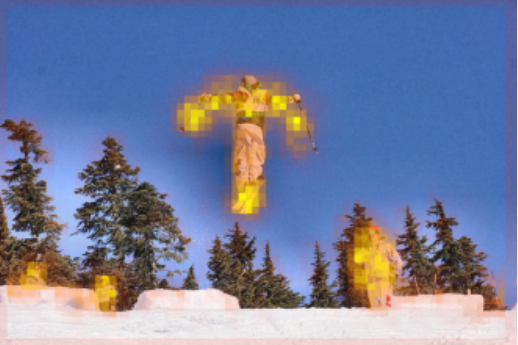}
  }\hspace{-6pt}
  \subfloat{
    \includegraphics[height=0.105\textwidth]{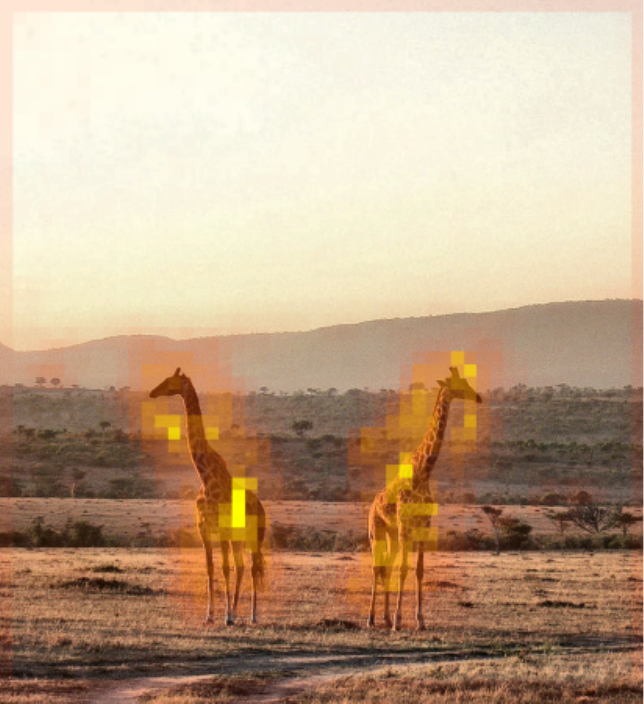}
  }
  \caption{Visualization of salience confidence at multi-scale feature maps on MSSD, CSD and COCO.}
  \label{fig:visualization of salience confidence of multi-scale features}
\end{figure}

\textbf{Salience Confidence.} \Cref{fig:visualization of salience confidence of multi-scale features} visualizes the salience confidence on MSSD, CSD, and COCO 2017 datasets. The visualization demonstrates that salience guides defect regions to achieve high confidence even for those with small sizes. 
Additionally, the confidence of large-sized objects decreases from the center to the border, benefiting fine-grained supervision. 
Interestingly, the salience labels are constructed solely based on bounding box annotations; however, the predicted confidence can further match rough object contours. This suggests the possibility that salience supervision may also benefit pixel-level tasks, such as instance segmentation. Therefore, transitioning from instance-level annotations to pixel-level predictions based on salience is a promising direction for future research.

\begin{figure}[htbp]
  \centering
  \includegraphics[width=0.45\textwidth]{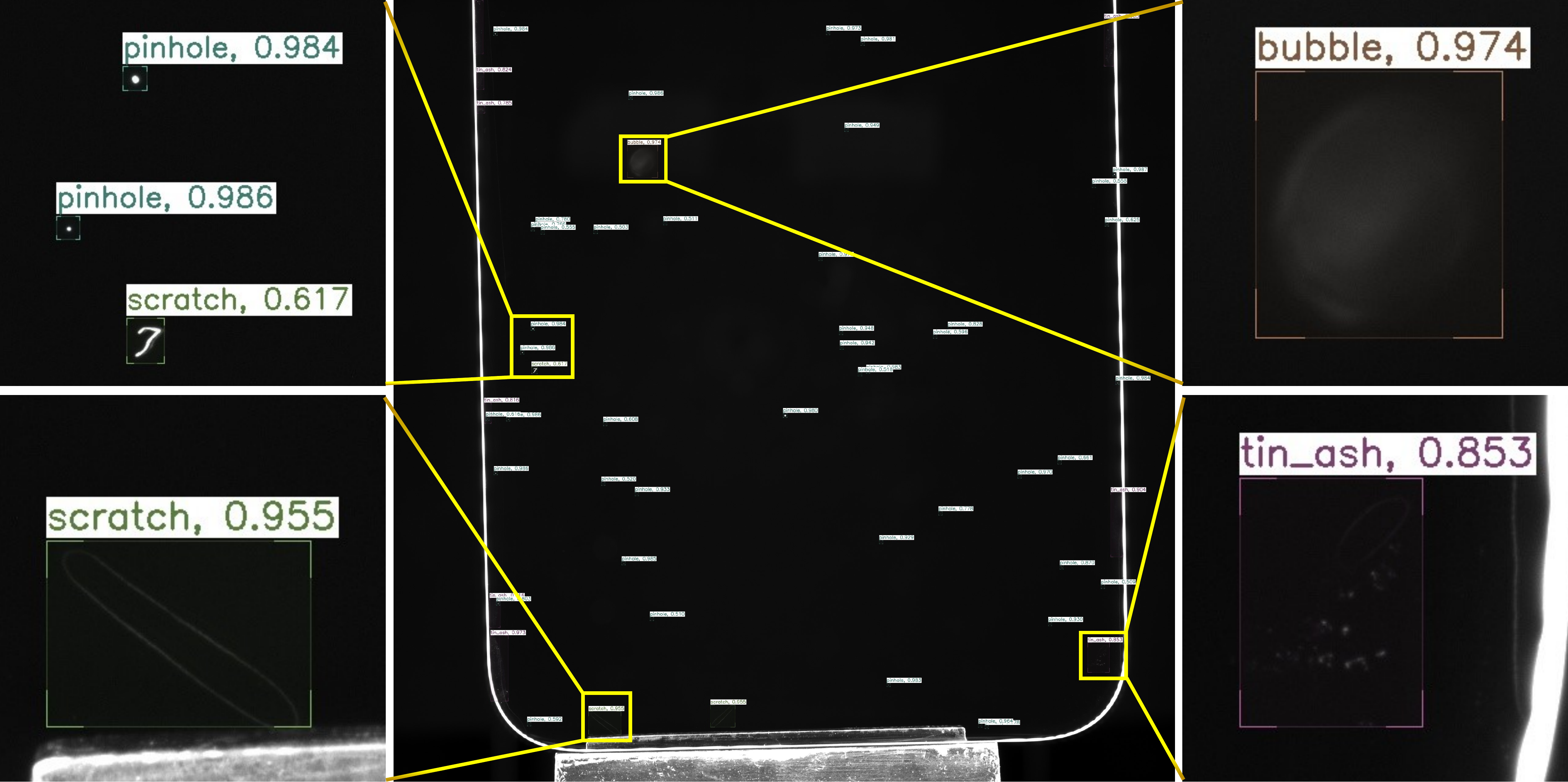}
  \caption{Visualization of detection results on the MSSD dataset.}
  \label{fig:visualization of detection results on the mssd dataset}
\end{figure}

\textbf{Detection results.} As shown in \Cref{fig:visualization of detection results on the mssd dataset}, the detection results of Salience DETR illustrate a strong adaptability to various appearances of objects, including 
small size (\eg pinhole), indiscriminative contours (\eg tin\_ash), intra-class differences (\eg scratch), and clear objects (\eg bubble). Moreover, Salience DETR shows a high level of confidence in distinguishing between categories, which enables it to achieve stable detection for confusable objects. 
\subsection{Inference cost}
\Cref{tab:inference time and memory} shows that Salience DETR without cross-level token fusion achieves comparable performance with other approaches at a lower computational cost. Further, introducing the module leads to better performance, with reasonable parameter and memory increase. These results show that Salience DETR obtains a good trade-off between performance and inference cost.
\begin{table}[htbp]
  \small
  \centering
  \setlength{\tabcolsep}{1.7pt}
  \caption{Inference time and memory}
  \label{tab:inference time and memory}
  \begin{tabular}{c|cccc}
    \toprule
    Methods & AP(CSD)$\uparrow$ & infer time$\downarrow$ & \#parameter$\downarrow$ & memory$\downarrow$ \\ \midrule
    H-Def-DETR & 53.0 & 0.0834 & 44.4M & 201MB\\
    DINO & 53.0 & 0.0860 & 46M & 207MB\\
    Focus DETR & 52.3 & 0.0822 & 44.4M & 201MB\\
    Salience DETR$^*$ & 52.8 & 0.0769 & 46M & 209MB \\
    Salience DETR & 53.2 & 0.0819 & 56.1M & 249MB\\
    \bottomrule
  \end{tabular}
  \begin{tablenotes}[flushleft]
    \item * denotes a faster variant without the cross-level token fusion.
  \end{tablenotes}
\end{table}

\section{Conclusion}
\label{sec:conclusion}
The paper develops a transformer-based detection framework, \ie Salience DETR, to mitigate the encoding and selection redundancies in two-stage DETR-like detectors. The key component of Salience DETR, namely hierarchical salience filtering refinement, selectively encodes a fraction of discriminative queries under the supervision of scale-independent salience to overcome scale bias. By conducting background embedding and cross-level token fusion, our model effectively tackles the problem of semantic misalignment among queries at different levels and layers. Furthermore, it leverages an elaborate redundancy removal module to stabilize two-stage initialization.
We demonstrate that Salience DETR achieves state-of-the-art performance on three task-specific and one generic object detection datasets, as well as a superior balance between computation and precision. We believe that our work could facilitate future researches on query redundancy in DETR-like methods.
\section*{Acknowledgments}
This work was supported by the National Key Research and Development Program of China under Grant 2020AAA0108302, the National Natural Science Foundation of China under Grant 62327808, and the Fundamental Research Funds for Xi'an Jiaotong University under Grant xtr072022001.

{
    \small
    \bibliographystyle{ieeenat_fullname}
    \bibliography{main}
}

\end{document}